\documentclass[runningheads]{llncs}
\usepackage[T1]{fontenc}
\usepackage{graphicx}

\usepackage{tikz}
\usetikzlibrary{arrows.meta}
\usetikzlibrary{decorations.markings}
\usepackage{xcolor}
\usepackage{amsmath}
\usepackage{amsfonts}
\usepackage{mathtools}
\usepackage{booktabs}
\usepackage{tabularx}
\newcolumntype{C}{>{\centering\arraybackslash}X}
\usepackage{pgf}
\usepackage{caption}
\usepackage{svg}
\usepackage{pdfpages}
\usepackage{subcaption}
\usepackage[colorlinks=False, hidelinks]{hyperref}

\definecolor{TU_green}{HTML}{639A00}
\usepackage{blindtext}

\newcommand\va{\boldsymbol{a}}
\newcommand\ve{\boldsymbol{e}}
\newcommand\vi{\boldsymbol{i}}
\newcommand\vs{\boldsymbol{s}}
\newcommand\vu{\boldsymbol{u}}
\newcommand\vx{\boldsymbol{x}}

\newcommand\vtheta{\boldsymbol{\theta}}

\newcommand\vTheta{\boldsymbol{\mathrm{\Theta}}}

\newcommand\ru{\boldsymbol{\mathrm{u}}}
\newcommand\rw{\boldsymbol{\mathrm{w}}}

\newcommand\calA{\mathcal{A}}
\newcommand\calU{\mathcal{U}}
\newcommand\calS{\mathcal{S}}
\newcommand\calT{\mathcal{T}}
\newcommand\calX{\mathcal{X}}

\newcommand\sE{\mathbb{E}}
\newcommand\sN{\mathbb{N}}
\newcommand\sR{\mathbb{R}}
\newcommand\sW{\mathbb{W}}

\begin{document}

\title{Optimizing Operation Recipes with Reinforcement Learning for Safe and Interpretable Control of Chemical Processes}

\titlerunning{Optimizing Operation Recipes with Reinforcement Learning}

\author{Dean Brandner
\and
Sergio Lucia
}

\authorrunning{D. Brandner and S. Lucia}

\institute{TU Dortmund University,
Dortmund, 
Germany \\
\email{\{dean.brandner, sergio.lucia\}@tu-dortmund.de}
}

\maketitle

\begin{abstract}
Optimal operation of chemical processes is vital for energy, resource, and cost savings in chemical engineering.  The problem of optimal operation can be tackled with reinforcement learning, but traditional reinforcement learning methods face challenges due to hard constraints related to quality and safety that must be strictly satisfied, and the large amount of required training data. Chemical processes often cannot provide sufficient experimental data, and while detailed dynamic models can be an alternative, their complexity makes it computationally intractable to generate the needed data. Optimal control methods, such as model predictive control, also struggle with the complexity of the underlying dynamic models. Consequently, many chemical processes rely on manually defined operation recipes combined with simple linear controllers, leading to suboptimal performance and limited flexibility.

In this work, we propose a novel approach that leverages expert knowledge embedded in operation recipes. By using reinforcement learning to optimize the parameters of these recipes and their underlying linear controllers, we achieve an optimized operation recipe. This method requires significantly less data, handles constraints more effectively, and is more interpretable than traditional reinforcement learning methods due to the structured nature of the recipes. We demonstrate the potential of our approach through simulation results of an industrial batch polymerization reactor, showing that it can approach the performance of optimal controllers while addressing the limitations of existing methods.
\end{abstract}

\section{Introduction}
The chemical industry is the largest industrial energy consumer and the third largest industrial emitter of CO2 after the steel and cement industries, making it necessary to achieve high efficiencies together with innovative technologies and recycling to enable achieve net zero emissions \cite{iea2023tracking}. At the same time, chemical processes need to be very carefully operated so that strict quality, safety and regulatory requirements are fulfilled.

The optimal operation of chemical processes can be formulated as an optimal control problem or a Markov decision process (MDP) for which reinforcement learning (RL) has been recently explored \cite{nian2020review,shin2019reinforcement}. However, traditional RL techniques struggle with the consideration of hard constraints and need a large amount of data. Unfortunately, in the field of chemical engineering, many hard constraints related to quality and safety requirements need to be strictly ensured~\cite{arendtEvaluatingProcessSafety2000} and obtaining a large amount of experimental data for training is typically not possible.
The latter challenge can be alleviated by using detailed dynamic models of the chemical processes instead of interacting with the real plant itself but these models are often very complex, making it computationally infeasible to generate large amounts of data. Finally, chemical processes are typically still operated or supervised by humans, for which an interpretable operation strategy is beneficial.

A more established approach to perform advanced operation of chemical processes is the use of optimal control theory methods such as nonlinear model predictive control (NMPC) \cite{rawlingsModelPredictiveControl2020}. In this approach, a dynamic system model is used to obtain predictions and an optimal trajectory of control inputs is calculated by solving an optimization problem every time a new control input needs to be computed. NMPC has been successfully applied in many domains since it can directly deal with nonlinear multivariable systems with hard constraints. However, when the underlying dynamic models of the process are very complex, including for example partial differential equations, multi-phase systems or startup behavior of different unit operations, the resulting optimization problems are often intractable. While some approaches exist to alleviate this problem, such as tailored fast optimization solvers~\cite{verschueren2022acados}, the use of approximate MPC based on neural networks~(NN) \cite{chen2018approximating,karg2020efficient} or the combination of RL and NMPC \cite{zanon2020safe,brandner2024reinforced}, it remains challenging to solve the resulting optimization problems in real time.

As a result, even nowadays batch processes are mostly controlled in the following hierarchical fashion.
In the upper layer, a reference trajectory of setpoints, called the operation recipe, is provided. These recipes can either be rigorously calculated, or derived by an expert via trial-and-error.
The lower layer attempts tracking of the recipe references during execution of a batch run. Usually, simple linear PID controllers are used to track these references. 
Lots of research was put into optimizing these operation recipes in the past. However, the approaches either focus on bias correction of empirical models once a full batch is completed, such as in run-to-run control~\cite{campbellComparisonRuntorunControl2002}, or model-based trajectory optimization between or even during runs~\cite{kimRobustBatchtoBatchOptimization2019,chenParticleFiltersState2005}.
Although these model-based approaches lead to improvements, they require a detailed control oriented model. In practise, these kinds of models are often not available, inexact, or extremely difficult, if not impossible to use in model-based optimization.
Further, model-free optimization approaches such as RL can also be applied to find optimized recipes. Different approaches, ranging from application of standard RL techniques for batch recipe optimization, to newly custom-made RL methods are reviewed in~\cite{yooReinforcementLearningBatch2021}.
Still, the authors of~\cite{yooReinforcementLearningBatch2021} identify that data efficiency and constraint handling remain an issue for RL.
Due to the practical inapplicability of these approaches, batch processes are mostly controlled according to manually tuned operation recipes, which are the result of a combination of the experience of experts and heuristics~\cite{Recipes_BrandRihm23,Startup_for_reactive_distillation_Reepmeyer2004}.
The deployed reference trajectories are often constrained to ramps or constant holding signals, both applied until a certain condition is met.
The recipe parameters such as the slope of the ramps or the constant value are usually only tuned by experts and not by rigorous optimization. Further, also the tracking PID controllers must be tuned according to the recipe parameters. All this clearly leads to a significant suboptimal performance of batch processes.

In this work, we propose a new method to incorporate the expert knowledge embedded in operation recipes and combine it with the capabilities of RL when used with detailed dynamic models of complex chemical processes. We use a RL agent to optimally tune the parameters of the operation recipes as well as the parameters of the underlying linear controllers. The goal is to significantly increase the performance of operation recipes, approaching the optimal control solution which typically cannot be computed in real time.
Since the amount of deployed actions, which take the form of recipe and PID parameters, to run a full batch is small compared to traditional direct RL techniques, we argue that it is significantly easier to train and also easier to obtain a policy that satisfies hard constrains.
In addition, the resulting strategy is easily interpretable, as it retains the structure of operation recipes and linear controllers that is typical in chemical engineering.
We showcase the potential of the approach with simulation results of an industrial semi-batch polymerization reactor. This example can serve as a benchmark from chemical engineering for other methodologies, as it is a challenging system with strongly nonlinear dynamics, multiple inputs and several hard constraints for which traditional RL techniques struggle to find a suitable policy.

\section{Background}
\subsection{Reinforcement Learning}
RL aims at solving MDPs~\cite{suttonReinforcementLearningIntroduction2018}.
An MDP is composed of an agent and an environment.
At each time instance, the environment is in state~$\vs \in \calS \subseteq \sR^{n_{\vs}}$ and receives an action~$\va \in \calA \subseteq \sR^{n_{\va}}$ that is calculated according to the agent's policy~$\pi$. The sets $\calS$ and $\calA$ denote the sets of possible states and actions.
The policy can either be stochastic~$\va \sim \pi(\cdot \vert \vs)$, so a mapping from a state to a probability distribution over the action space, or deterministic~$\va = \pi(\vs)$, so a direct mapping from a state to a specific action.
For ease of notation, we will focus on deterministic policies for the rest of this work. However, all presented concepts also work with stochastic policies. 
When action~$\va$ is applied to the environment, the environment transitions from the current state~$\vs$ to the subsequent state~$\vs^+ \in \calS$ according to its underlying transition probability~$p(\vs^+ \vert \vs, \va)$, leading to:
\begin{align}
    \vs^+ \sim p(\cdot \vert \vs, \va) \label{eq:StateTransitionProbabilityDist}.
\end{align}
Often times, the transition from $\left< \vs, \va\right>$ to $\vs^+$ can also be expressed as a dynamic system model~$f: \calS \times \calA \times \sW \rightarrow \calS$ in which the uncertainty of~\eqref{eq:StateTransitionProbabilityDist} is accounted for via the random variable~$\rw \sim \sW$ leading to:
\begin{align}
    \vs^+ = f(\vs, \va, \rw). \label{eq:StateTransitionDynamicModel}
\end{align}
In the case of a deterministic environment, the random variable is always zero~$\rw = 0$ and the transition probability $p$ becomes the Dirac impulse.
After the environment transitions one step, the agent receives the next state~$\vs^+$ and a scalar reward~$r\in\sR$, which measures how good the state-action-pair $\left< \vs, \va \right>$ is according to a previously designed objective.
After that, the cycle of providing an action~$\va$ given state~$\vs$ and observing the subsequent state~$\vs^+$ and reward~$r$ is repeated.
The overall goal of RL is to find the optimal policy~$\pi^\star$ that maximizes the expected cumulative reward~$J(\pi) \in \sR$ based on the interaction of the agent with the environment only, by collecting the transition to the subsequent states~$\vs^+$ and rewards~$r(\vs, \va)$ when being in state~$\vs$ and taking action~$\va$.
Once the optimal policy~$\pi^\star$ is found, the MDP is assumed to be solved.
To calculate the expected cumulative reward~$J(\pi)$, we introduce the state value function~$V^{\pi}: \calS \rightarrow \sR$.
The state value function is recursively defined in~\eqref{eq:State-Value-Function} as the sum of the immediate reward~$r(\vs, \va)$ when being at state~$\vs$ and taking action~$\va$ according to policy~$\pi$, and the expected value of the state value~$V^{\pi}(\vs^+)$ over all possible subsequent states~$\vs^+$ according to the transition probability~$p$, that is:
\begin{align}
    V^{\pi}(\vs) = r(\vs, \va)\vert_{\va = \pi(\vs)}
    + \gamma \sE_{\vs^+
    }
    [V^{\pi}(\vs^+)]. \label{eq:State-Value-Function}
\end{align}
where $0 < \gamma \leq 1$ is a discount factor.
By taking the expected value over the initial states~$\vs_0$, the expected cumulative reward~$J(\pi)$ can be derived as 
\begin{align}
    J(\pi) = \sE_{\vs_0} [V^{\pi} (\vs) ].
\end{align}
A policy is optimal when it maximizes the expected cumulative reward
\begin{align}
    \pi^\star = \arg \max_{\pi} J(\pi). \label{eq:OptimalPolicy}
\end{align}
However, the maximization problem in~\eqref{eq:OptimalPolicy} requires to solve an infinite dimensional optimization problem which is intractable in general.
The most common solution to this problem is to approximate the optimal policy~$\pi_{\vtheta^\star}(\vs) \approx \pi^\star$ with a function approximator with parameters~$\vtheta \in \sR^{n_{\vtheta}}$.
Due to its universal approximation capabilities~\cite{UniversalApproximation_Hornik_1989}, often NNs are used.
The maximization problem boils down to finding the optimal parameters~$\vtheta^\star$
\begin{align}
    \vtheta^\star = \arg \max_{\vtheta} J(\vtheta). \label{eq:OptimalPolicyParam}
\end{align}

RL gives a rich toolbox of algorithms to solve~\eqref{eq:OptimalPolicyParam}.
Among others, common approaches try to find the optimal policy by directly updating the policy parameters~$\vtheta$ with the policy gradient~$\nabla_{\vtheta} J(\vtheta) \in \sR^{n_{\vtheta}}$ according to the gradient-ascent optimization algorithm
\begin{align}
    \vtheta \leftarrow \vtheta + \alpha \nabla_{\vtheta} J(\vtheta).
\end{align}
Since the calculation of the policy gradient~$\nabla_{\vtheta} J(\vtheta)$ from the interaction of the agent with the environment is not straightforward, many algorithms exist that address this task.
For deterministic policies, state-of-the-art performance can be achieved among others with the twin delayed deep deterministic policy gradient~(TD3) algorithm~\cite{TD3_Fujimoto18a} while proximal policy optimization~(PPO)~\cite{PPO_Schulman17} and soft-actor-critic~(SAC)~\cite{SAC_Haarnoja18b} are state-of-the-art for stochastic polices.

Although serious advances were made in recent years, the sample efficiency of RL algorithms is still poor and scales badly with increasingly complex tasks, as well as with the state and action space dimension. This is especially the case for safe RL. An extensive overview is given in~\cite{guReviewSafeReinforcement2024a}.
In the context of process control, the consideration of hard constraints remains a major challenge. Although constrained RL methods exist, they are mostly accounted for via penalty terms in the reward function (that is, as soft constraints) in practice.

\subsection{Standard and Advanced Control Approaches}
\subsubsection{Linear PID Control}
The majority of chemical processes is operated in steady state or at least a pseudo steady state, which is equivalent to having only slowly changing global process dynamics.
Since often the nonlinear system dynamics can be approximated sufficiently with a linear model close around a steady state, and also due to its easy practical implementation, most controllers in chemical industry are proportional-integral-differential~(PID) controllers~\cite{skogestadMultivariableFeedbackControl2005}.

We want to highlight the differences between the RL state~$\vs$ and RL action~$\va$, and their physical counterparts. Therefore, we introduce the physical dynamic system state~$\vx\in \calX \subseteq \sR^{n_{\vx}}$ and the physical control input~$\vu\in \calU \subseteq \sR^{n_{\vu}}$.
The discretized dynamics of the physical system are described by the potentially nonlinear model~$f_{\mathrm{p,d}}: \calX \times \calU \times \sW \rightarrow \calX$  that can also be stochastic, accounted for by the random variable~$\rw$
\begin{align}
    \vx_{k+1} = f_{\mathrm{p,d}}(\vx_k, \vu_k, \rw_k) \label{eq:DynamicPhysicalSystem_Nonlinear_discrete}
\end{align}
with~$k$ being the current sampling time.
We want to emphasize that the RL state and action can in fact be the physical state and control input, so $\vs = \vx$ and $\va = \vu$, but the RL state and RL action can also augment further effects such as states from past time steps~$\vx_{k-1}$ or past control actions~$\vu_{\mathrm{prev}}$.

Given a desired steady state~$\vx_{\mathrm{ss}}$ and~$\vu_{\mathrm{ss}}$ of~\eqref{eq:DynamicPhysicalSystem_Nonlinear_discrete}, PID controllers try to minimize the error~$\ve \in \sR^{n_{\vx}}$ between the state~$\vx$ and the desired steady state~$\vx_{\mathrm{ss}}$
\begin{align}
    \ve = \vx - \vx_{\mathrm{ss}}.
\end{align}
To achieve this, the applied control input~$\vu$ is calculated according to
\begin{align}
    \vu_k = \vu_\mathrm{ss} + K_{\mathrm{P}} \, \ve_k + K_{\mathrm{I}} \, \sum \ve_k \, \mathrm{\Delta} t + K_{\mathrm{D}} \, \dot{\ve}_k. \label{eq:PIC_control}
\end{align}
The control input~$\vu_k$ is therefore determined by adapting the steady state input~$\vu_{\mathrm{ss}}$ with three different correction terms.
The displayed correction terms reflect a correction due to the immediate error~$\ve_k$, the integrated error~$\sum \ve_k \, \mathrm{\Delta} t$ and the differentiated error~$\dot{\ve}_k$.
Each correction term is multiplied by a controller gain~$K_i$, which must be tuned by an expert to achieve the desired performance.
Most chemical processes are controlled with PI controllers only, which is done by setting~$K_{\mathrm{D}} = 0$.
The full PID parameter vector~$\vTheta_\mathrm{PID} \in \calT_\mathrm{PID}$ includes all parameters that influence the controller performance, such as the setpoints~$\vx_\mathrm{ss}$ and the controller gains~$K_i$.

Cascade control is a strategy that utilizes an inner and an outer control loop to control a system with two subsystems exhibiting differently fast dynamics.
The outer, slower control loop determines the setpoint for the inner, faster control loop, which then rapidly adjusts to follow this setpoint set by the outer loop~\cite{skogestadMultivariableFeedbackControl2005}.
Cascade control is frequently used in the chemical industry, such as in controlling the reactor temperature in chemical reactors.
In this scenario, the outer loop controls the reactor temperature by providing setpoints for the jacket temperature, which is controlled by the inner loop.

Although PID controllers are widely used in chemical industry, they have limitations when applied to highly nonlinear systems without a steady state in the desired operating range.
The theory behind PID controllers assumes linear dynamics, making it difficult to handle high nonlinearity.
Additionally, process constraints, like maximum or minimum reactor temperatures, can only be addressed indirectly through proper controller tuning.
Online consideration of these constraints is not possible.
The performance of PID controllers heavily depends on their tuning.
Poorly tuned controllers result in unsatisfactory control performance.
Finding reasonable controller gains can be a cumbersome task.

\subsubsection{Nonlinear Model Predictive Control}
NMPC is an advanced control approach that is used frequently in chemical engineering, and addresses the shortcomings of linear PID control such as rigorous constraint consideration and optimized performance.
To account for both, NMPC solves the following optimal control problem each sampling time~\cite{rawlingsModelPredictiveControl2020}:
\begin{subequations}
    \label{eq:NMPC_problem}
    \begin{align}
        \ru^\star = \arg \min_{\ru} &\quad F_{\mathrm{f}}(\vx_N) + \sum_{k = 0}^{N-1} \ell(\vx_k, \vu_k) \label{eq:NMPC_objective}\\
        \mathrm{s.t.} & \quad \vx_{k+1} = f_\mathrm{p,d}(\vx_k, \vu_k), \quad \vx_0 = \vx(t_k) \label{eq:NMPC_systemeq},\\
        & \quad g(\vx_k, \vu_k) \leq 0. \label{eq:NMPC_process_constraints}
    \end{align}
\end{subequations}
Within this optimization problem, the system states~$\vx_k$ are internally simulated according to~\eqref{eq:NMPC_systemeq} for a prediction horizon of~$N$ steps starting from the most recently measured system state~$\vx(t_k)$.
On this considered prediction horizon, process constraints~$g:\calX \times \calU \rightarrow \sR^{n_{g}}$ in the form of~\eqref{eq:NMPC_process_constraints} are evaluated at each sampling instance.
The performance is then optimized by finding the optimal control input sequence~$\ru^\star = [\vu_0^\star, \ldots, \vu_{N-1}^\star]^\top$ according to the objective function~\eqref{eq:NMPC_objective}, which includes the stage cost~$\ell(\vx_k, \vu_k)$, and the terminal cost~$F_\mathrm{f}(\vx_N)$, which compensates the truncation errors due to a finite prediction horizon.
The first element~$\vu_0^\star$ of the optimal control sequence~$\ru^\star$ is then applied to the system.
Note that in classical control theory, typically minimization problems are solved, while RL aims at maximizing the reward.
Thus, the stage cost in NMPC can be interpreted as the negative reward in RL.

Despite alleviating many problems from linear PID control and performing optimally on constrained systems, the performance of NMPC can significantly deteriorate with model errors. 
To avoid model errors, very detailed models can be formulated, which in turn can lead to very complex optimization problems due to highly nonlinear dynamics or a large amount of optimization variables preventing its solution in real-time.  

\subsection{Heuristics and Operation Recipes}
In control of chemical systems, the operation of batch processes is a major challenge, despite its omnipresence in the production of pharmaceuticals and special chemicals.
Advanced model-based optimal control approaches can often not be applied due to the high model complexity and the potential loss of real-time applicability.
Batch operation is often done by a hierarchical approach, which is composed of an upper recipe layer, which provides setpoints for the lower tracking layer with mostly simple linear PID controllers.
The reference trajectories of the recipe layer can be calculated by model-based optimization, provided that a good model exists that can be used for optimization. Since this is rarely the case due to high complexity, optimization cannot be performed. Hence, these recipes are designed by experts and are therefore mostly constrained to simple patterns.

In this contribution, we will align the definition of an operation recipe, which is tailored by expert knowledge, to the definition used in~\cite{Recipes_BrandRihm23}.
An operation recipe is a sequential procedure that separates the whole batch cycle into smaller batch phases~$z = 1, \ldots,n_z$.
As an illustrative example, one can consider a tank that is filled until a certain level is reached in the first phase~$z=1$, operated at the desired level in the second phase~$z=2$, and emptied in the third phase~$z=3$.
These phases themselves are also made up out of smaller sub-steps~$c = 1, \ldots, n_c$, which are the decisions that are applied to the system.
In general, batch phase~$z$ can only be completed if the final step~$c_z$ of the phase~$z$ is reached.
The procedure in each sub-step~$c$ is determined by a qualitative decision, like setting some value of an actuator or waiting until a certain condition is met.
The values that are assigned by these manual adaptations are part of the recipe parameters~$\vTheta_\mathrm{R} \in \calT_\mathrm{R}$.
The total set of the parameters for the whole batch cycle are thus the combination of the recipe and PID parameters~$\vTheta^\top = [\vTheta_\mathrm{R}^\top, \vTheta_\mathrm{PID}^\top] \in \calT = \calT_\mathrm{R} \times \calT_\mathrm{PID}$.
The quantitative value that is set to the qualitative operation in step~$c$ can then be read from the~$c$-th element of the full parameter vector~~$\vTheta_c \in \calT_c \subseteq \sR$.
In classic operation recipes, these values are either set once or can be adapted by an expert according to the current plant situation.
Table~\ref{tab:ExampleRecipe} summarizes the concept of operation recipes at the example of the considered tank above.
\begin{table}[!t]
    \centering
    \caption{Example recipe for filling and emptying a tank.} \label{tab:ExampleRecipe}
    \begin{tabularx}{\textwidth}{c c c l}
        \toprule
         \hspace{0.25cm} Phase~$z$ \hspace{0.25cm} & \hspace{0.25cm} Step~$c$ \hspace{0.25cm} & \hspace{0.25cm} Type \hspace{0.25cm} & Description  \\ \midrule\midrule 
         1 & 1 & set & Start the feed pump to a feed rate of $\vTheta_{1}$\\
         1 & 2 & condition & Wait until the level reaches $\vTheta_{2}$\\
         2 & 3 & set & Stop the feed pump and wait for a time of $\vTheta_{3}$\\
         3 & 4 & set & Open the outlet valve to a percentage of $\vTheta_{4}$\\
         3 & 5 & condition & Wait until the level falls below $\vTheta_{5}$\\
         \bottomrule
    \end{tabularx}
\end{table}

While this approach has the advantage that the decisions are made by an expert and that the operation recipe is highly interpretable, this approach leads typically to conservative results and is sensitive to PID controller tuning.

\section{Proposed Approach: Recipe-based Reinforcement Learning}

\begin{figure}[!b]
    \begin{minipage}{0.48\textwidth}
        % Direct RL
        \begin{subfigure}[b]{\linewidth}
            \centering
            \begin{tikzpicture}[
                Green_rounded_rectangle_node/.style={rounded corners, draw=TU_green!90, fill=TU_green!25, minimum width = 4 cm, minimum height = 1 cm, align = center},
                ]
                % RL
                % Controller/Agent
                \filldraw[rounded corners, draw = TU_green!80, fill = TU_green!30] (0, 0) rectangle (4, 1);
                \node at (2, 1 - 0.25) {RL agent};
                \node at (2, 1 - 0.25 - 0.5) {$\va = \vu = \pi_{\vtheta}(\vs)$};
    
                % Environment
                \filldraw[rounded corners, draw = TU_green!80, fill = TU_green!30] (0, -0.5) rectangle (4, -0.5 - 1);
                \node at (2, -0.5 - 0.25) {System};
                \node at (2, -0.5 - 0.25 - 0.5) {$\vs = \vx_{k+1} = f_\mathrm{p,d}(\vx_k, \vu_k)$};
    
                % Arrows
                \draw[<-] (1.5, 0) -- node[left] {$\vx$} (1.5, -0.5);
                \draw[->] (2.5, 0) -- node[right] {$\vu$} (2.5, -0.5);
            \end{tikzpicture}
            \caption{Direct RL} \label{subfig:DirectRL}
        \end{subfigure}

        \bigskip
        % NMPC
        \begin{subfigure}[b]{\linewidth}
            \centering
            \begin{tikzpicture}[
                Green_rounded_rectangle_node/.style={rounded corners, draw=TU_green!90, fill=TU_green!25, minimum width = 4 cm, minimum height = 1 cm, align = center},
                ]
                % NMPC
                % Controller/Agent
                \filldraw[rounded corners, draw = TU_green!80, fill = TU_green!30] (0, 0) rectangle (4, 1);
                \node at (2, 1 - 0.25) {NMPC};
                \node at (2, 1 - 0.25 - 0.5) {$\vu_0^\star = \mathrm{solve~\eqref{eq:NMPC_problem}}$};
    
                % Environment
                \filldraw[rounded corners, draw = TU_green!80, fill = TU_green!30] (0, -0.5) rectangle (4, -0.5 - 1);
                \node at (2, -0.5 - 0.25) {System};
                \node at (2, -0.5 - 0.25 - 0.5) {$\vx_{k+1} = f_\mathrm{p,d}(\vx_k, \vu_k)$};
    
                % Arrows
                \draw[<-] (1.5, 0) -- node[left] {$\vx$} (1.5, -0.5);
                \draw[->] (2.5, 0) -- node[right] {$\vu$} (2.5, -0.5);
            \end{tikzpicture}
            \caption{NMPC} \label{subfig:NMPC}
        \end{subfigure}
        % \hspace{0.01\textwidth
        
        \bigskip
        \centering
        \begin{subfigure}[b]{\linewidth}
        \centering
            \begin{tikzpicture}[
                Green_rounded_rectangle_node/.style={rounded corners, draw=TU_green!90, fill=TU_green!25, minimum width = 4 cm, minimum height = 1 cm, align = center},
                ]
                % Recipes and PID
                \filldraw[rounded corners, draw = TU_green!80, fill = TU_green!20] (0, 0) rectangle (4, 0 + 0.25 + 1 + 1);
                \filldraw[rounded corners, draw = TU_green!80, fill = TU_green!30] (0 + 0.25, 0 + 0.25) rectangle (4 - 0.25, 0 + 0.25 + 1);
                \node at (2, 0 + 0.25 + 1 - 0.25) {PID};
                \node at (2, 0 + 0.25 + 1 - 0.25 - 0.5) {$\vu_{\mathrm{PID}} = \mathrm{from~\eqref{eq:PIC_control}}$};
                \node at (2, 0 + 0.25 + 1 + 1 - 0.25) {Recipe with $\vTheta = \mathrm{const.}$};
                \node at (2, 0 + 0.25 + 1 + 1 - 0.25 - 0.5) {$\vu_{\mathrm{R}}= \mathrm{from~Table~\ref{tab:ExampleRecipe}}$};
    
                % Environment
                \filldraw[rounded corners, draw = TU_green!80, fill = TU_green!30] (0, -0.5) rectangle (4, -0.5 - 1);
                \node at (2, -0.5 - 0.25) {System};
                \node at (2, -0.5 - 0.25 - 0.5) {$\vx_{k+1} = f_\mathrm{p,d}(\vx_k, \vu_k)$};
    
                % Arrows
                \draw[<-] (1.5, 0) -- node[left] {$\vx$} (1.5, -0.5);
                \draw[->] (2.5, 0) -- node[right] {$\vu$} (2.5, -0.5);
            \end{tikzpicture}
            \caption{Recipes and PID ($\vTheta = \mathrm{const.}$)} \label{subfig:RecipeAndPID}
        \end{subfigure}
    \end{minipage}
    \hfill
    \begin{subfigure}[c]{0.48\textwidth}
        \centering
        \begin{tikzpicture}[
            Green_rounded_rectangle_node/.style={rounded corners, draw=TU_green!90, fill=TU_green!25, minimum width = 4 cm, minimum height = 1 cm, align = center},
            ]
            % RL
            % Controller/Agent
            \filldraw[rounded corners, draw = TU_green!80, fill = TU_green!30] (0, 0) rectangle (4, 1);
            \node at (2, 1 - 0.25) {RL agent};
            \node at (2, 1 - 0.25 - 0.5) {$\va = \vTheta_c = \pi_{\vtheta}(\vs)$};

            % Environment
            \filldraw[rounded corners, draw = TU_green!80, fill = TU_green!10] (0, -1) rectangle (4, -1 - 1 -1 -1 -1 -1 - 0.25 - 0.1);
            \node at (2, -1 - 0.25) {Environment};
            \node at (2, -1 - 0.25 - 0.5) {$\vs^\top = [\vx^\top, \vTheta^\top, c]$};

            \filldraw[rounded corners, draw = TU_green!80, fill = TU_green!20] (0 + 0.25, -1 - 1 - 0.1) rectangle (4 - 0.25, -1 - 1 - 1 - 1 - 0.25 - 0.1);
            \node at (2, -1 - 1 - 0.25 - 0.1) {Recipe with $\vTheta_{\mathrm{R}}$};
            \node at (2,  -1 - 1 - 0.25 - 0.5 - 0.1) {$\vu_{\mathrm{R}}= \mathrm{from~Table~\ref{tab:ExampleRecipe}}$};

            \filldraw[rounded corners, draw = TU_green!80, fill = TU_green!30] (0 + 0.25 + 0.25, -1 - 1 - 1 - 0.1) rectangle (4 - 0.25 - 0.25, -1 - 1 - 1 - 1 - 0.1);
            \node at (2, -1 - 1 - 1 - 0.25 - 0.1) {PID  with $\vTheta_{\mathrm{PID}}$};
            \node at (2, -1 - 1 - 1 - 0.25 - 0.5 - 0.1) {$\vu_{\mathrm{PID}} = \mathrm{from~\eqref{eq:PIC_control}}$};

            \filldraw[rounded corners, draw = TU_green!80, fill = TU_green!30] (0.25, -1 - 1 - 1 - 1 - 1 - 0.1) rectangle (4 - 0.25, -1 - 1 - 1 - 1 - 1 - 1 - 0.1);
            \node at (2, -1 - 1 - 1 - 1 - 1 - 0.25 - 0.1) {System};
            \node at (2, -1 - 1 - 1 - 1 - 1 - 0.25 - 0.5 - 0.1) {$\vx_{k+1} = f_\mathrm{p,d}(\vx_k, \vu_k)$};

            % % Arrows
            \draw[<-] (1.5, -1 - 1 - 1 - 1 - 0.25 - 0.1) -- node[left] {$\vx$} (1.5, -1 - 1 - 1 - 1 - 1 - 0.1);
            \draw[->] (2.5, -1 - 1 - 1 - 1 - 0.25 - 0.1) -- node[right] {$\vu$} (2.5, -1 - 1 - 1 - 1 - 1 - 0.1);

            \draw[<-] (1.5, 0) -- node[left] {$\vs$} (1.5, -1);
            \draw[->] (2.5, 0) -- node[right] {$\va$} (2.5, -1);
        \end{tikzpicture}
        \caption{Proposed approach: Recipe and PID parameters via RL agent
        }
    \end{subfigure}
    \caption{Established control approaches (left) vs. proposed approach (right).} \label{fig:comparison_of_concepts}
\end{figure}
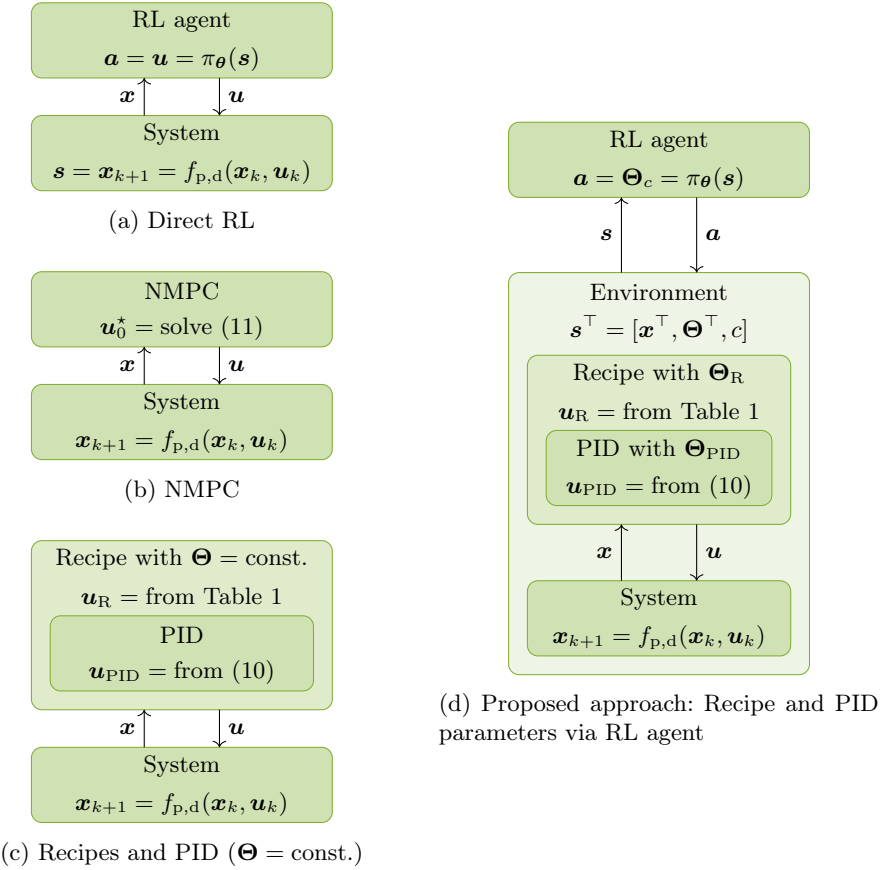

In our proposed approach, we train an RL agent with a NN policy that receives the current RL state of the system and computes the next optimal recipe and PID parameters.
Contrary to the classical implementation of expert-tuned operation recipes and PID controllers, in which the parameters are typically fixed once and not adapted afterwards, our approach delivers the optimal parameters depending on the current physical system state allowing an optimized control performance.
While classical RL aims at finding the optimal control policy directly, which should in theory result in a similar performance as NMPC, it often struggles to find a reasonable policy when considering complex systems with hard constraints.
The same problem also occurs when the optimized trajectory is calculated in advance and tracked as in the hierarchical batch operation setting.
In addition, the resulting policy is usually not interpretable because the computed control action does not give any information about future decisions, and also process constraints are not considered explicitly as in NMPC. 
In contrast, our approach adapts the parameters of a fixed operation recipe structure and the parameters of the PID controllers.
This results in a highly structured policy so even in the beginning of the RL training process, the resulting policies have an acceptable performance, leading to improved learning behavior.
Also, both the operation recipe and the PID controllers were originally designed by experts and should therefore be operationally safe within a certain expert-certified parameter space.
Lastly, since the parameters are part of the operation recipe, the derived policy is easy to interpret.
Figure~\ref{fig:comparison_of_concepts} summarizes all presented control approaches (left column) and contrasts them with our proposed approach (right column).

In the classical RL setting (see Figure~\ref{subfig:DirectRL}), the RL state~$\vs_\mathrm{cl}$ and the RL action~$\va_\mathrm{cl}$ are usually the physical state~$\vs_\mathrm{cl} = \vx$ and the physical control input~$\va_\mathrm{cl} = \vu$, or at least closely related to it.
The agent therefore learns the optimal policy~$\va_\mathrm{cl} = \pi_{\vtheta}(\vs_\mathrm{cl})$ directly.
The underlying environment dynamics are governed by~\eqref{eq:DynamicPhysicalSystem_Nonlinear_discrete}.
The reward~$r_\mathrm{cl}$ typically depends on a single transition of the physical system only.
Based on this information, the optimal policy is to be derived.

We propose a different design of the environment, specialized for learning recipe and PID parameters.
Instead of only the physical system, the environment consists also of the operation recipe and PID structure.
We define the RL state~$\vs_\mathrm{R} \in \calX \times \calT \times \sN$
as the combination of the physical state~$\vx$, the recipe and PID parameters~$\vTheta$ and the current recipe step~$c$
\begin{align}
     \vs^\top_\mathrm{R} = [ \vx^\top, \vTheta^\top, c ]
    .
\end{align}
We will assume that the parameter vector~$\vTheta$ is ordered, meaning that the $c$-th element in the parameter vector corresponds to the $c$-th recipe step.
The RL action~$\va_\mathrm{R}$ reduces from the full physical control input~$\vu$ to the $c$-th parameter~$\va_\mathrm{R} = \vTheta_c$ that is required in the $c$-th recipe step.
The agent therefore learns to predict the best recipe or PID parameter with its policy~$\va_\mathrm{R} = \pi_{\vtheta} (\vs_\mathrm{R})$.
The dynamics of the RL state~$\vs_\mathrm{R}$ now have to be adapted.
Since in each interaction of the agent with the environment, the parameter vector~$\vTheta$ changes due to the applied action~$\va_\mathrm{R} = \vTheta_c$, the dynamics of the parameter vector~$\vTheta$ are given as
\begin{align}
    \vTheta^+ = \vTheta + \vi_c \, \vTheta_c \label{eq:R_Param_update}
\end{align}
with~$\vi_c$ being the standard unit vector in the $c$-direction.
Further, as the $c$-th parameter is set, the step counter~$c$ must be increased
\begin{align}
    c^+ = c + 1.
\end{align}
Since the physical system now either does not transition at all, which is the case when not all parameters of a certain batch phase~$z$ are set~($c \neq c_z$), or the physical system transitions through a whole batch phase~$z$, which is the case when all parameters in batch phase~$z$ are set~($c = c_z$), the transition dynamics must be adapted accordingly.
We introduce the dynamic system~$\hat{f}_\mathrm{p,d}$, which models the transition through a whole recipe step~$z$.
This is equivalent to sequentially evaluating the dynamic system~\eqref{eq:DynamicPhysicalSystem_Nonlinear_discrete} with the control actions~$\vu$ according to recipe phase~$z$ until the criterion to switch from phase~$z$ to $z+1$ is met.
The overall transition dynamics of the physical system can be summarized as
\begin{align}
    \vx^+ = \begin{cases}
        \vx & \mathrm{if} ~ c \neq c_z, \\
        \hat{f}_\mathrm{p,d}(\vx, \vu) & \mathrm{else}.
    \end{cases}
\end{align}

Finally, also the reward~$r_\mathrm{R}$ must be adapted accordingly.
If only the parameters are changed according to~\eqref{eq:R_Param_update}, the physical system does not change
and the reward is always zero for those transitions.
However, if a full batch phase is carried out~($c = c_z$), the classical reward~$r_\mathrm{cl}$ can be evaluated at each time instance until the end of the batch phase after $n_\mathrm{end}(\vs_\mathrm{R})$ transitions.
The resulting reward~$r_\mathrm{R}$ is then summed together and weighted with the discretization time step~$\mathrm{\Delta} t$.
The calculation can be summarized as
\begin{align}
    r_\mathrm{R}(\vs_\mathrm{R}, \va_\mathrm{R}) = \begin{dcases}
        0 & \mathrm{if}~c \neq c_z,\\
        \sum_{i=k}^{n_{\mathrm{end}} (\vs_\mathrm{R})} r_\mathrm{cl}(\vx_i, \vu_i) \, \mathrm{\Delta} t & \mathrm{else}.
    \end{dcases}
\end{align}

\section{Experiments}
\begin{figure}[!b]
    \centering
    \includegraphics[width = 0.9\textwidth]{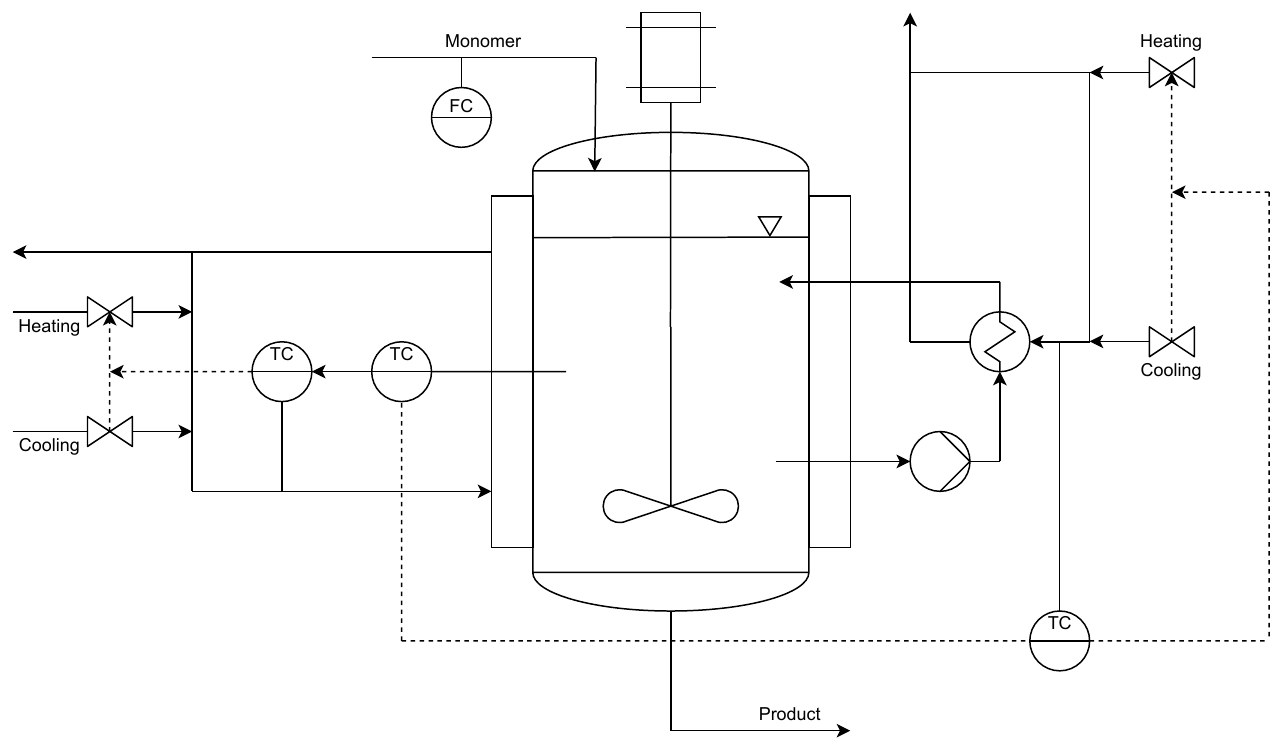}
    \caption{Sketch of the polymerization reactor.} \label{fig:PolyReactor}
\end{figure}

We investigate the proposed approach with the example of a semi-batch polymerization reactor~\cite{lucia2014handling}.
Figure~\ref{fig:PolyReactor} shows a sketch of the reactor.
The reactor content consists three components: water, monomer and product.
Their masses are given as~$m_\mathrm{W}$, $m_\mathrm{M}$ and $m_\mathrm{P}$ respectively.
The reactor content with a temperature~$T_\mathrm{R}$ is in direct contact with the walls of temperature~$T_\mathrm{S}$.
The reactor contains a jacket with a temperature~$T_\mathrm{J}$, and an external heat exchanger with temperature~$T_\mathrm{EHE}$.
Both can be used for heating and cooling of the reactor content.
Further, the temperature of the water on the cooling side of the external heat exchanger is given as~$T_\mathrm{CW,EHE}$.
The reactor can be filled via the feed stream~$\dot{m}_\mathrm{feed}$ that consists of monomer and water.
The jacket temperature~$T_\mathrm{J}$ and the temperature of the external heat exchanger~$T_\mathrm{EHE}$ can be controlled by the inlet water temperature to these devices, which are~$T_\mathrm{J,in}$ and~$T_\mathrm{CW,EHE,in}$.
Lastly, although being no real physical states, we also model the accumulated feed mass~$m_\mathrm{acc}$ and the adiabatic temperature~$T_\mathrm{ad}$ as for both process constraints exist.
Hence, the resulting states~$\vx$ and control inputs~$\vu$ are
\begin{align}
    \vx &= [m_\mathrm{W}, m_\mathrm{M}, m_\mathrm{P}, T_\mathrm{R}, T_\mathrm{S}, T_\mathrm{J}, T_\mathrm{EHE}, T_\mathrm{CW,EHE}, m_\mathrm{acc}, T_\mathrm{ad}]^\top,\\
    \vu &= [m_\mathrm{feed}, T_\mathrm{J,in}, T_\mathrm{CW,EHE,in}]^\top.
\end{align}
The governing equations and parameter values can be found in~\cite{lucia2014handling} and online\footnote{\label{ftnote:dompc}\url{www.do-mpc.com}}.

A classical recipe-based control approach, which is inspired from the closed-loop trajectory of economic NMPC, is shown in Table~\ref{tab:PolyRktRecipe}.
The whole batch process can be divided into~$n_z = 3$ batch phases with 14 parameters~$\vTheta$ in total.
In the first two phases, the feed stream~$\dot{m}_\mathrm{feed}$ is ramped up until a certain value is reached.
After that, the feed stream~$\dot{m}_\mathrm{feed}$ is kept at a constant rate until the whole batch cycle terminates.
During all three batch phases, the reactor temperature~$T_\mathrm{R}$ is controlled via PID controllers in a cascade control structure.
The outer slower controller tracks the reference temperature~$T_\mathrm{R,ref}$ by calculating setpoints for the jacket temperature~$T_\mathrm{J,ref}$ and the temperature of the external heat exchanger~$T_\mathrm{EHE,set}$, which are both each controlled by a much faster inner controller.
All differential gains~$K_\mathrm{D}$ of the PID controllers are set to zero.
\begin{table}[!tb]
    \centering
    \caption{Parameterized operation recipe of the polymerization reactor.} \label{tab:PolyRktRecipe}
    \begin{tabularx}{\textwidth}{c c c l}
        \toprule
         Phase~$z$ \hspace{0.10cm} & Step~$c$ \hspace{0.10cm} & Type \hspace{0.1cm} & Description  \\ \midrule\midrule 
         1 & 1 & set & Set slope of feed stream ramp to $\vTheta_{1}$\\
         1 & 2 & set & Set $T_\mathrm{R,set}$ of outer PID controller to $\vTheta_{2}$\\
         1 & 3 & set & Set $K_\mathrm{P}$ of outer PID controller to $\vTheta_{3}$\\
         1 & 4 & set & Set $K_\mathrm{I}$ of outer PID controller to $\vTheta_{4}$\\
         1 & 5 & condition & Run phase~1 until total mass reaches $\vTheta_{5}$\\
         2 & 6 & set &  Set slope of feed stream ramp to $\vTheta_{6}$\\
         2 & 7 & set &  Set $T_\mathrm{R,set}$ of outer PID controller to $\vTheta_{7}$\\
         2 & 8 & set & Set $K_\mathrm{P}$ of outer PID controller to $\vTheta_{8}$\\
         2 & 9 & set & Set $K_\mathrm{I}$ of the outer PID controller to $\vTheta_{9}$\\
         2 & 10 & set &  Set maximal allowed value of feed stream to $\vTheta_{10}$\\
         2 & 11 & condition & Run phase~2 until elapsed times is larger than~$\vTheta_{11}$\\
           &    & &    or feed stream becomes larger than $\vTheta_{10}$ \\
         3 & 12 & set & Close feed and set $T_\mathrm{R,set}$ of outer PID controller to $\vTheta_{12}$\\
         3 & 13 & set & Set $K_\mathrm{P}$ of outer PID controller to $\vTheta_{13}$\\
         3 & 14 & set & Set $K_\mathrm{I}$ of outer PID controller to $\vTheta_{14}$\\
         \bottomrule
    \end{tabularx}
\end{table}

Optimal control of the polymerization reactor usually has the goal to produce most product in the shortest amount of time, while satisfying process constraints.
Often, also a smooth control trajectory is preferred to avoid damage to the actuators.
Since the considered polymerization reaction always results in full conversion of the monomer to the product, the batch cycle is assumed to be finished when 99\,\% of all possible reactable monomer is reacted.
When setting up an optimization problems for this batch cycle, it turns out that maximizing product mass and minimizing the batch time result in the same solution.
Solving time-optimal control problems is challenging in general, which is the reason why mostly product maximization is performed in practise.
However, RL can theoretically deal with both objectives.
We investigate our approach on three different learning scenarios
\begin{enumerate}
    \item Maximize product mass~$m_\mathrm{P}$,
    \item Minimize batch time~$t_\mathrm{batch}$,
    \item Minimize batch time~$t_\mathrm{batch}$ and maximize product mass~$m_\mathrm{P}$ (hybrid).
\end{enumerate}
The designed reward therefore encodes the objectives above.
In addition to that, all rewards also encode that the optimal control policy shall not violate constraints and should be reasonable smooth.
For this, constraint violations are penalized with a high cost, and a reasonably smooth policy is penalized by taking large steps in the physical control input. 
As it is a special case for operation recipes, in which PID controllers are tuned, all rewards also penalize error between the reactor temperature~$T_\mathrm{R}$ and the setpoint~$T_\mathrm{R,set}$.

\begin{figure}[!t]
    \centering
    \resizebox{\columnwidth}{!}{
        \input{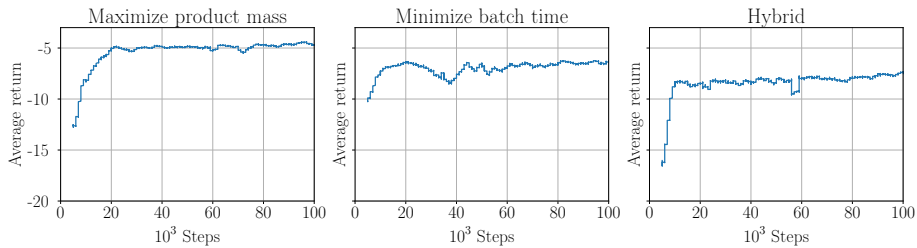}
    }
    \caption{Learning curves of the RL agents for all three different scenarios.
    }
    \label{fig:learning_curves}
\end{figure}

For all three scenarios, a hyperparameter gridsearch with all possible 96 combinations from Table~\ref{tab:Gridsearch} is carried out.
The policy and Q-function are approximated with feedforward NNs with ReLU activation functions.
All other hyperparameters remain at the default values of the stable-baselines3~\cite{stable-baselines3} implementation.
We observe that almost all agents converge to a good or at least reasonable policy.
Still, there is room for improvement.
Figure~\ref{fig:learning_curves} show the learning curves for the best agents trained for each scenario.
The agents learn rapidly in the beginning and start to converge after $\mathrm{40\cdot10^3}$ iterations at the latest and only improve marginally afterwards.
Fastest convergence is achieved by the hybrid reward scenario, which is expected as it carries most externally provided extra information in the form of two non-conflicting objectives.
Note that the absolute value of the return does not provide information on the policy quality as it encodes different information.
\begin{table}[!t]
    \centering
    \caption{Considered hyperparameters for gridsearch.}\label{tab:Gridsearch}
    \begin{tabular}{cc|cc} \toprule
          \hspace{0.25cm} Hyperparameter \hspace{0.25cm} & \hspace{0.5cm} Range  \hspace{0.5cm} & \hspace{0.25cm} Hyperparameter \hspace{0.25cm} & \hspace{0.5cm} Range \hspace{0.5cm} \\ \midrule\midrule
          RL Algorithm & SAC, TD3 & Policy architecture & [50, 50], [50, 25, 10]\\
         Batch size & 512, 4\,096 & Learning rate & $\mathrm{3\cdot 10^{-4}}$, $10^{-5}$ \\
         Exploration noise & 0, 0.1 & Buffer size & $10^6$, $10^5$, $10^4$ \\\bottomrule
    \end{tabular}
\end{table}

All trained agents are evaluated with respect to their common performance metrics, which are the average batch time~$\bar{t}_\mathrm{batch}$ and the averaged absolute and relative number of constraint violations~$\bar{n}_\mathrm{CV}$ and $\bar{n}_\mathrm{CV, rel}$.
To evaluate the average batch time~$\bar{t}_\mathrm{batch}$, the agents control the system from 50 initial conditions, which were sampled from a different seed than the training was performed on.
The metrics for all initial conditions are measured and averaged.
We compare the performance of the three agents to NMPC~(see Figure~\ref{subfig:NMPC}), which uses the exact system model and a large prediction horizon of $N=30$ with a discretization interval of 30\,s.
This NMPC deals as an estimate of the optimal policy and provides a benchmark performance.
Furthermore, we compare the agents to a non-adaptive recipe with reasonable fixed parameters~(see Figure~\ref{subfig:RecipeAndPID}).
This fixed recipe deals as a baseline.
Lastly, we also try to compare our method to direct RL~(see Figure~\ref{subfig:DirectRL}) to evaluate our method at a reference.
Like for the recipe RL agents, we trained the direct RL agents for all three scenarios and performed a  hyperparameter gridsearch, considering all possible combinations from Table~\ref{tab:Gridsearch}.
From all 288 investigated agents, only 58 terminated all 50 batches, while the remaining agents did not achieve 99\,\% conversion within five hours and were truncated consequently.
From the remaining 58 agents, only two agents violated constraints in less than 5\,\% of all observed states.
The agent with lowest percentage of constraint violations is considered the best and used as a comparison.

The results are illustrated in~Table~\ref{tab:ResultsBatchtime}.
As expected, the NMPC delivers the shortest average batch time and can therefore be referenced as the benchmark.
The manually tuned baseline recipe results in an average performance.
Since all direct RL approaches struggled during training due to stability and overall convergence, even the best obtained agent has a larger average batch time than the baseline recipe. Further, more constraint violations than the baseline can be observed. This illustrates that despite serious tuning effort, tuning of the environment and the RL algorithm can be a cumbersome task. 
On the other hand, from all three investigated recipe-based training scenarios, scenario~3 (hybrid) shows the best control performance, which is likely as the most expert knowledge is put in the design of the reward.
This is congruent with the best learning speed as shown in Figure~\ref{fig:learning_curves}.
Still, all scenarios resulted with a final policy that outperforms both, the baseline recipe and direct RL, and appears to be similar in all cases.
Also, all investigated scenarios showed a good learning performance and had no stability issues as in direct RL. We argue that this behavior originates from to the structured recipe environment.
All resulting operation recipes are in average more than 1\,h faster than the manually tuned baseline recipe.
Further, all recipe-based RL agents do not violate constraints on the investigated batches, while the direct RL agent violates the constraints in 1.54\,\% of all states.
This also highlights that constraining the RL agent to the structure of operation recipes can improve the overall safety of the RL agent.
\begin{table}[!tb]
    \centering
    \caption{Performance evaluation of the different approaches and scenarios.}\label{tab:ResultsBatchtime}
    \tabularx{\textwidth}{@{\extracolsep{\stretch{1}}}*{7}{c}@{}} 
        \toprule
        Method & $\bar{t}_\mathrm{batch} /\mathrm{[h]}$ & $\bar{n}_\mathrm{CV} / [-]$ & $\bar{n}_\mathrm{CV,rel} / [\mathrm{\%}]$ \\ \midrule \midrule
        Baseline (Recipe and PID) &  3.29 $\pm$ 0.07 & 0.14 $\pm$ 0.49 & 0.03 $\pm$ 0.12 \\ 
        Benchmark (NMPC) & 1.37 $\pm$ 0.04 & 0 & 0 \\
        Reference (Direct RL) & 3.42 $\pm$ 0.04 & 6.32 $\pm$ 1.97 & 1.54 $\pm$ 0.52 \\
        Scenario~1 (Maximize $m_\mathrm{P}$) & 2.28 $\pm$ 0.10 & 0 & 0\\
        Scenario~2 (Minimize $t_\mathrm{batch}$) & 2.25 $\pm$ 0.05 & 0 & 0 \\
        Scenario~3 (Hybrid) & 2.21 $\pm$ 0.06 & 0 & 0\\\bottomrule
\end{table}

The code with all experiments is available online\footnote{\url{https://github.com/DeanBrandner/Optimizing_Recipes_with_RL.git}}.
All RL training runs are performed with the toolbox stable-baselines3~\cite{stable-baselines3} and CasADi~\cite{Andersson2019}.
The results for NMPC are obtained with the toolbox do-mpc~\cite{fiedler2023mpc}.

\section{Conclusion}
This paper proposes a novel approach for the optimal operation of chemical processes by integrating reinforcement learning with expert knowledge encapsulated in structured operation recipes. This method addresses key challenges in chemical engineering, such as handling hard constraints related to quality and safety, and the limited availability of experimental data for training. By optimizing the recipe and controller parameters, our approach achieves near-optimal performance with significantly less data and improved constraint handling.

The simulation results of an industrial batch polymerization reactor illustrate the effectiveness of our method. Compared to traditional reinforcement learning and model predictive control, our approach offers enhanced interpretability, leveraging the structured knowledge of operation recipes. 

Future work will focus on extending this methodology to integrate further expert knowledge via reinforcement learning with human feedback. Additionally, real-world implementation and validation of our approach will be performed to confirm its practical viability and benefits in industrial settings.
Also, larger case studies will be investigated to assess the scalability of the proposed approach.

\begin{credits}
\subsubsection{\ackname}
This work was funded by the Deutsche Forschungsgemeinschaft (DFG, German Research Foundation) – 466380688 – within the Priority Program “SPP 2331: Machine Learning in Chemical Engineering”.
\end{credits}
\bibliographystyle{splncs04}
\bibliography{bibliography.bib}

@misc{iea2023tracking,
  title        = {{Tracking Clean Energy Progress 2023}},
  author       = {{IEA}},
  year         = 2023,
  note         = {Licence: CC BY 4.0},
  url          = {https://www.iea.org/reports/tracking-clean-energy-progress-2023},
  institution  = {IEA},
  address      = {Paris}
}

@article{nian2020review,
  title={A review on reinforcement learning: Introduction and applications in industrial process control},
  author={Nian, Rui and Liu, Jinfeng and Huang, Biao},
  journal={Computers \& Chemical Engineering},
  volume={139},
  pages={106886},
  year={2020},
  publisher={Elsevier}
}

@article{shin2019reinforcement,
  title={Reinforcement learning--overview of recent progress and implications for process control},
  author={Shin, Joohyun and Badgwell, Thomas A and Liu, Kuang-Hung and Lee, Jay H},
  journal={Computers \& Chemical Engineering},
  volume={127},
  pages={282--294},
  year={2019},
  publisher={Elsevier}
}

@misc{PPO_Schulman17,
  title = {Proximal {{Policy Optimization Algorithms}}},
  author = {Schulman, John and Wolski, Filip and Dhariwal, Prafulla and Radford, Alec and Klimov, Oleg},
  year = {2017},
  month = aug,
  number = {arXiv:1707.06347},
  eprint = {1707.06347},
  publisher = {arXiv},
  urldate = {2024-02-05},
  archiveprefix = {arXiv},
}

@book{rawlingsModelPredictiveControl2020,
  title = {Model Predictive Control: Theory, Computation, and Design},
  shorttitle = {Model Predictive Control},
  author = {Rawlings, James Blake and Mayne, David Q. and Diehl, Moritz},
  year = {2020},
  edition = {2nd},
  publisher = {Nob Hill Publishing},
  address = {Santa Barbara, California},
  isbn = {978-0-9759377-5-4},
  lccn = {TJ217.6 .R39 2020}
}

@article{Recipes_BrandRihm23,
  title = {Efficient Dynamic Sampling of Batch Processes through Operation Recipes},
  author = {Brand Rihm, Gerardo and Esche, Erik and Repke, Jens-Uwe},
  year = {2023},
  journal = {Computers \& Chemical Engineering},
  volume = {179},
  pages = {108433},
  issn = {00981354},
  langid = {english},
}

@inproceedings{SAC_Haarnoja18b,
  title = {Soft Actor-Critic: {{Off-policy}} Maximum Entropy Deep Reinforcement Learning with a Stochastic Actor},
  booktitle = {Proceedings of the 35th International Conference on Machine Learning},
  author = {Haarnoja, Tuomas and Zhou, Aurick and Abbeel, Pieter and Levine, Sergey},
  editor = {Dy, Jennifer and Krause, Andreas},
  year = {2018-07-10/2018-07-15},
  series = {Proceedings of Machine Learning Research},
  volume = {80},
  pages = {1861--1870},
  publisher = {PMLR},
}

@book{skogestadMultivariableFeedbackControl2005,
  title = {Multivariable Feedback Control: Analysis and Design},
  shorttitle = {Multivariable Feedback Control},
  author = {Skogestad, Sigurd and Postlethwaite, Ian},
  year = {2005},
  edition = {2nd ed},
  publisher = {John Wiley},
  address = {Hoboken, NJ},
  isbn = {978-0-470-01167-6},
  lccn = {TJ216 .S47 2005},
}

@book{suttonReinforcementLearningIntroduction2018,
  title = {Reinforcement Learning: An Introduction},
  shorttitle = {Reinforcement Learning},
  author = {Sutton, Richard S. and Barto, Andrew G.},
  year = {2018},
  series = {Adaptive Computation and Machine Learning Series},
  edition = {2nd},
  publisher = {The MIT Press},
  address = {Cambridge, Massachusetts},
  isbn = {978-0-262-03924-6},
  lccn = {Q325.6 .R45 2018},
}

@inproceedings{TD3_Fujimoto18a,
  title = {Addressing Function Approximation Error in Actor-Critic Methods},
  booktitle = {Proceedings of the 35th International Conference on Machine Learning},
  author = {Fujimoto, Scott and {van Hoof}, Herke and Meger, David},
  editor = {Dy, Jennifer and Krause, Andreas},
  year = {2018-07-10/2018-07-15},
  series = {Proceedings of Machine Learning Research},
  volume = {80},
  pages = {1587--1596},
  publisher = {PMLR},
}

@article{verschueren2022acados,
  title={acados—a modular open-source framework for fast embedded optimal control},
  author={Verschueren, Robin and Frison, Gianluca and Kouzoupis, Dimitris and Frey, Jonathan and Duijkeren, Niels van and Zanelli, Andrea and Novoselnik, Branimir and Albin, Thivaharan and Quirynen, Rien and Diehl, Moritz},
  journal={Mathematical Programming Computation},
  volume={14},
  number={1},
  pages={147--183},
  year={2022},
  publisher={Springer}
}

@article{karg2020efficient,
  title={Efficient representation and approximation of model predictive control laws via deep learning},
  author={Karg, Benjamin and Lucia, Sergio},
  journal={IEEE Transactions on Cybernetics},
  volume={50},
  number={9},
  pages={3866--3878},
  year={2020},
  publisher={IEEE}
}

@INPROCEEDINGS{brandner2024reinforced,
  author={Brandner, Dean and Lucia, Sergio},
  booktitle={2024 European Control Conference (ECC)}, 
  title={Reinforced Model Predictive Control via Trust-Region Quasi-Newton Policy Optimization}, 
  year={2024},
  volume={},
  number={},
  pages={2299-2305},
}

@article{zanon2020safe,
  title={Safe reinforcement learning using robust MPC},
  author={Zanon, Mario and Gros, S{\'e}bastien},
  journal={IEEE Transactions on Automatic Control},
  volume={66},
  number={8},
  pages={3638--3652},
  year={2020},
  publisher={IEEE}
}

@inproceedings{chen2018approximating,
  title={Approximating explicit model predictive control using constrained neural networks},
  author={Chen, Steven and Saulnier, Kelsey and Atanasov, Nikolay and Lee, Daniel D and Kumar, Vijay and Pappas, George J and Morari, Manfred},
  booktitle={2018 Annual American control conference (ACC)},
  pages={1520--1527},
  year={2018},
  organization={IEEE}
}

@article{fiedler2023mpc,
  title={do-mpc: Towards FAIR nonlinear and robust model predictive control},
  author={Fiedler, Felix and Karg, Benjamin and L{\"u}ken, Lukas and Brandner, Dean and Heinlein, Moritz and Brabender, Felix and Lucia, Sergio},
  journal={Control Engineering Practice},
  volume={140},
  pages={105676},
  year={2023},
  publisher={Elsevier}
}

@article{lucia2014handling,
  title={Handling uncertainty in economic nonlinear model predictive control: A comparative case study},
  author={Lucia, Sergio and Andersson, Joel AE and Brandt, Heiko and Diehl, Moritz and Engell, Sebastian},
  journal={Journal of Process Control},
  volume={24},
  number={8},
  pages={1247--1259},
  year={2014},
  publisher={Elsevier}
}

@article{stable-baselines3,
  author  = {Antonin Raffin and Ashley Hill and Adam Gleave and Anssi Kanervisto and Maximilian Ernestus and Noah Dormann},
  title   = {Stable-Baselines3: Reliable Reinforcement Learning Implementations},
  journal = {Journal of Machine Learning Research},
  year    = {2021},
  volume  = {22},
  number  = {268},
  pages   = {1-8},
}

@Article{Andersson2019,
  author = {Joel A E Andersson and Joris Gillis and Greg Horn
            and James B Rawlings and Moritz Diehl},
  title = {{CasADi} -- {A} software framework for nonlinear optimization
           and optimal control},
  journal = {Mathematical Programming Computation},
  volume = {11},
  number = {1},
  pages = {1--36},
  year = {2019},
  publisher = {Springer},
}

@article{Startup_for_reactive_distillation_Reepmeyer2004,
title = {Time optimal start-up strategies for reactive distillation columns},
journal = {Chemical Engineering Science},
volume = {59},
number = {20},
pages = {4339-4347},
year = {2004},
issn = {0009-2509},
author = {Frauke Reepmeyer and Jens-Uwe Repke and Günter Wozny},
}

@article{UniversalApproximation_Hornik_1989,
title = {Multilayer feedforward networks are universal approximators},
journal = {Neural Networks},
volume = {2},
number = {5},
pages = {359-366},
year = {1989},
issn = {0893-6080},
author = {Kurt Hornik and Maxwell Stinchcombe and Halbert White},
}

@book{arendtEvaluatingProcessSafety2000,
  title = {Evaluating Process Safety in the Chemical Industry: A User's Guide to Quantitative Risk Analysis},
  shorttitle = {Evaluating Process Safety in the Chemical Industry},
  author = {Arendt, J. S. and Lorenzo, D. K.},
  year = {2000},
  series = {A {{CCPS}} Concept Book},
  publisher = {American Chemistry Council ; Center for Chemical Process Safety},
  address = {Arlington, Va. : New York},
  isbn = {978-0-8169-0746-5},
  lccn = {TP155.5 .A68 2000},
}

@misc{guReviewSafeReinforcement2024a,
  title = {A {{Review}} of {{Safe Reinforcement Learning}}: {{Methods}}, {{Theory}} and {{Applications}}},
  shorttitle = {A {{Review}} of {{Safe Reinforcement Learning}}},
  author = {Gu, Shangding and Yang, Long and Du, Yali and Chen, Guang and Walter, Florian and Wang, Jun and Knoll, Alois},
  year = {2024},
  month = may,
  number = {arXiv:2205.10330},
  eprint = {2205.10330},
  primaryclass = {cs},
  publisher = {arXiv},
  urldate = {2024-07-19},
  archiveprefix = {arXiv},
}

@article{yooReinforcementLearningBatch2021,
  title = {Reinforcement Learning for Batch Process Control: {{Review}} and Perspectives},
  shorttitle = {Reinforcement Learning for Batch Process Control},
  author = {Yoo, Haeun and Byun, Ha Eun and Han, Dongho and Lee, Jay H.},
  year = {2021},
  journal = {Annual Reviews in Control},
  volume = {52},
  pages = {108--119},
  issn = {13675788},
}

@inproceedings{campbellComparisonRuntorunControl2002,
  title = {A Comparison of Run-to-Run Control Algorithms},
  booktitle = {Proceedings of the 2002 {{American Control Conference}} ({{IEEE Cat}}. {{No}}.{{CH37301}})},
  author = {Campbell, W.J. and Firth, S.K. and Toprac, A.J. and Edgar, T.F.},
  year = {2002},
  pages = {2150-2155 vol.3},
  publisher = {IEEE},
  address = {Anchorage, AK, USA},
  isbn = {978-0-7803-7298-6},
}

@article{kimRobustBatchtoBatchOptimization2019,
  title = {Robust {{Batch-to-Batch Optimization}} with {{Scenario Adaptation}}},
  author = {Kim, Boeun and Huusom, Jakob K. and Lee, Jay H.},
  year = {2019},
  month = jul,
  journal = {Industrial \& Engineering Chemistry Research},
  volume = {58},
  number = {30},
  pages = {13664--13674},
  issn = {0888-5885, 1520-5045},
}

@article{chenParticleFiltersState2005,
  title = {Particle Filters for State and Parameter Estimation in Batch Processes},
  author = {Chen, Tao and Morris, Julian and Martin, Elaine},
  year = {2005},
  month = sep,
  journal = {Journal of Process Control},
  volume = {15},
  number = {6},
  pages = {665--673},
  issn = {09591524},
}
\end{document}